# Action Recognition based on Subdivision-Fusion Model

Zongbo Hao[1]
zbhao@uestc.edu.cn

Linlin Lu[1]
fountainous@gmail.com

Qianni Zhang[2]
qianni.zhang@qmul.ac.uk

Jie Wu[1]
jiewu.uestc@gmail.com

Ebroul Izquierdo[2]
ebroul.izquierdo@qmul.ac.uk

Juanyu Yang[1]
yangjuanyu2008@163.com

Jun Zhao[1]
zhaojun.hjh@gmail.com

[1] School of Information and
Software Engineering
University of Electronic Science
and Technology of China

[2] Multimedia and Vision Group
School of Electronic Engineering
and Computer Science
Queen Mary, University of London

**Abstract**

This paper proposes a novel Subdivision-Fusion Model (SFM) to recognize human actions. In most action recognition tasks, overlapping feature distribution is a common problem leading to overfitting. In the subdivision stage of the proposed SFM, samples in each category are clustered. Then, such samples are grouped into multiple more concentrated subcategories. Boundaries for the subcategories are easier to find and as consequence overfitting is avoided. In the subsequent fusion stage, the multi-subcategories classification results are converted back to the original category recognition problem. Two methods to determine the number of clusters are provided. The proposed model has been thoroughly tested with four popular datasets. In the Hollywood2 dataset, an accuracy of 79.4% is achieved, outperforming the state-of-the-art accuracy of 64.3%. The performance on the YouTube Action dataset has been improved from 75.8% to 82.5%, while considerably improvements are also observed on the KTH and UCF50 datasets.

# 1 Introduction

Automated action recognition has gained more and more interest in recent years [1], [3], [6], [25], [26], [27]. The key challenge is detecting, tracking and recognizing humans from the videos by computer vision, and then understanding and characterizing their actions [8]. Human action recognition has many applications including intelligent surveillance, content-based video retrieval, human-computer interaction and games design [17]. Because of individual differences, the diversity and complexity of actions, and complex backgrounds, action recognition is a daunting problem. The main challenges involved in this task are: (1) person localization in cluttered or dynamic environment; (2) adverse effects of lighting condition; and (3) fast moving dynamic background. At the same time, a robust human action recognition algorithm should be invariant to different rates of execution [5]. But for machine learning based methods, especially for the end-to-end approaches, such as deep learning, all these challenges become the problem of designing a proper network and extracting well performed features.

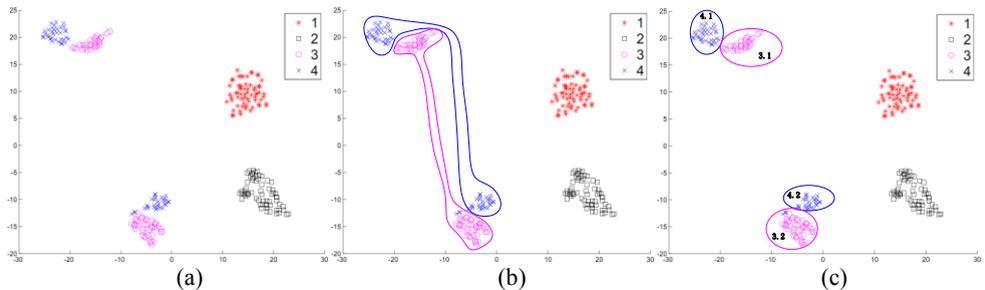

Figure 1: Subdivision in each class. (a) Sample distribution. Category 3 and 4 are both distributed in two regions and overlapped. (b) Finding boundaries to distinguish category 3 and 4 in a usual way, it is prone to be overfitting. (c) Clustering in each class makes it easier to distinguish the classes and avoid overfitting. In this case, the subclasses of 2 in class 3 and 2 in class 4, i.e. $K_3 = 2$ and $K_4 = 2$ will be considered as the proper $K_i$ parameters in our model, and class 1 and class 2 will not be subdivided, i.e. $K_1 = 1$ and $K_2 = 1$ (the related discussion is presented in section 3). This figure is best viewed in color.

There are two stages in automated human action recognition: action representation, and classification with these representations. Action representations encode the human actions, and have great effect on classification. An ideal representation should not only take the influence of the sizes of the human body, complex backgrounds, different viewpoints and the speed of the action into account, but also comprise sufficient information for the classifier to discriminate the actions. An effective classifier is expected to be able to distinguish existed and new action types [4].

In this paper, we propose a novel Subdivision-Fusion Model (SFM) to recognize human actions in the case of overlapping feature distribution. The paper is structured as follows. In section 2, the motivation and the model SFM is presented. In section 3, two methods to determine the proper number of clusters are provided. We evaluated SFM on four popular datasets to demonstrate the enhancement in performance in section 4. The paper is concluded in section 5.

## 2 Subdivision-Fusion Model (SFM)

### 2.1 Motivation

In most recognition tasks, the overlapping feature distribution is popular, as presented in Figure 1(a) (similar feature distribution can be referred to Figure 4(b), the 2-dimensional visualization of the high-dimensional feature space of KTH [20] dataset by t-Distributed Stochastic Neighbor Embedding (t-SNE) [13]). Category 3 and 4 are both distributed in two regions and overlapped, instead of distributing in separating regions as supposed, like category 1 and 2 do. As can be inspected in Figure 1(a), it may suggests two revelations: (1) those two categories have similar features; (2) both classes could be separated into two subclasses respectively, which may have quite meaningful distinction. In this case it is not wise to find the boundary that enclose each category into one region and not overlapped with others, as shown in Figure 1(b). Trying to find that kind of boundaries, overfitting will be obviously resulted in. Hence, category 3 and 4 can be divided into two subcategories respectively, and the smooth boundaries can be found as shown in Figure 1(c). The sample belongs to either purple red region, it belongs to category 3; and the sample belongs to either blue region, it belongs to category 4. In this paper, we propose a SFM model to solve this problem.

## 2.2 Model

Suppose there are $n$ samples $\{(x_1, y_1), \cdots, (x_n, y_n)\}, y_i \in \mathbb{R}^L, 1 \leq i \leq n$ in the dataset, which are grouped into $L$ categories, $x_i$ and $y_i$ denotes the data and the labels respectively. The features are extracted by some kind of models, for example a deep network:

$$F(X) = [f_1, f_2, ..., f_n] = h([x_1, x_2, ..., x_n]) \tag{1}$$

where $f_i$ is the feature vector for each sample $x_i$, $h$ is the transformation by the network from the sample space $\mathbb{R}^S$ to feature space $\mathbb{R}^F$. In the recognition task, a classifier such as SVM or Softmax is connected to the feature vector:

$$\psi(x_i) = [p_1, p_2, ..., p_L]^T \tag{2}$$

where $p_j$ is the category confidence value for category $j, 1 \leq j \leq L$. The accuracy depends on the network and the classifier. It is difficult to find the boundaries in the feature space in the case of overlapping, as shown in Figure 1(a). Motivated by section 2.1, we subdivide each category using clustering algorithm.

Assuming that the data of the $i^{th}$ category in the feature space can be grouped into $K_i$ clusters, the labels are updated by the results of clustering, all the samples in the same subcategory will have the same label. The samples will be with the new labels, $\{(x_1, y_1'), \cdots, (x_n, y_n')\}, y_i' \in \mathbb{R}^M$, where $M$ is the total number of updated categories, and $M = \sum_{i=1}^{L} K_i$. The output of the classifier will be:

$$\psi'(x_i) = [p_1, p_2, ..., p_M]^T \tag{3}$$

For the case in Figure 1(c), $K_1 = 1$, $K_2 = 1$, $K_3 = 2$, $K_4 = 2$, $L = 4$, $M = 6$. Please note that the clustering is made in each category respectively. By clustering and updating the labels, we made a transformation from the feature space to the sample space:

$$Y' = [y_1', y_2', ..., y_n'] = c([f_1, f_2, ..., f_n]) \tag{4}$$

where $c$ is the clustering operation, which transforms feature vectors to labels. With proper parameter $K_i$ (which will be discussed in section 3), the samples in the same subcategory will be more concentrated and the new smooth boundaries will be easier to find to avoid overfitting, as presented in Figure 1(c).

As for the deep network, we make the training again with the new labels and classifier to get the $M$-category classification result with the output of (3). In the following fusion stage, we add another layer to the classifier which combines the $M$-category output result of $\psi'(x_i)$ in (3) to the original $L$-category output result of $\psi(x_i)$ in (2), as presented in Figure 2.

As for the extracted or precomputed features, we update the label of each sample by the result of clustering, and then directly feed the updated features into to the classifier to get the *M*-category classification result with the output of (3). The fusion stage is similar to the operation in the deep network which is mentioned above, as presented in Figure 2.

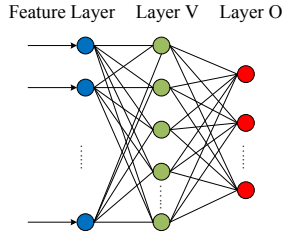

Figure 2: Classifier in the Fusion Stage

In Figure 2, Feature Layer is the output of the deep network or the precomputed features, in which each sample is represented as a feature vector. Layer V with *M* nodes acts as an *M*-category classifier. Then, we connect an output Layer O with *L* nodes to Layer V. Let $W(i,k)$ be the weight between node *k* in Layer V and node *i* in Layer O:

$$W(i,k) = \begin{cases} 1, V_k \in O_i \\ 0, V_k \notin O_i \end{cases} \quad (5)$$

where $V_k$ is the $k^{\text{th}}$ subcategory output in Layer V, $O_i$ is the $i^{\text{th}}$ category output in Layer O. If subcategory *k* belongs to category *i*, $W(i,k) = 1$, otherwise $W(i,k) = 0$. Thus, the final classification result *R* is calculated by:

$$R = \max(O_i), O_i = \sum_{k=1}^{M} W(i,k) * V_k, 1 \le i \le L, 1 \le k \le M \quad (6)$$

With Layer O, according to (6), the *M*-category classification is converted back to the *L*-category classification which is identical to the original problem.

Thus, our model improves the performance in two aspects as follows:

1) For the deep network, we train the given network twice, cascading by the process of updating the labels according to the results of clustering; we build a connection between the feature vectors and a new layer V, instead of directly connecting to the final output layer. In this way, the learned features can focus on more concentrated samples which are valuable for improving recognition accuracy; furthermore, the network is made deeper, leading to more invariable and abstract features.

2) For the extracted or precomputed features, it is easier for the classifier to find the boundaries with the data distributed in more centralized regions. Verma and Rahman [24] also pointed out that clustering the overlapping samples can improve the performance of the classifier.

The procedure of utilizing our model with the deep network and the extracted or precomputed features is presented in Figure 3(a) and Figure 3(b).

Sparse Subspace Clustering (SSC) [2] is employed in clustering in this paper. In contrast to the traditional k-means algorithm, SSC is able to tackle with both high-dimensional data and the noise of data efficiently.

The training process and testing process of our model are stated as follows:

**Algorithm1: Training Process**

**Input:** Labelled training data and a given feature extractor.

1) Train dataset of *L* categories by the given feature extractor and obtain feature vectors.

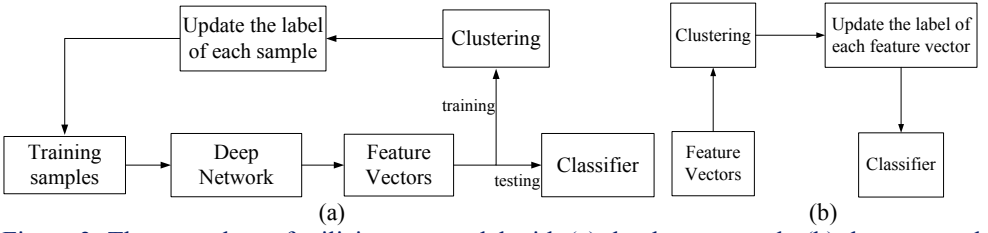

Figure 3: The procedure of utilizing our model with (a) the deep network, (b) the extracted or precomputed features.

2) Cluster in the feature space of each category by Sparse Subspace Clustering (SSC) [2] and get the fine-grained $M$ subcategories.

3) Update the labels of the training data according to the clustering results of step 2.

4) For the deep network, train the network again with the updated samples. For the extracted or precomputed features, train the classifier again.

**Output:** The updated recognition system with output nodes corresponding to $M$ subcategories.

**Algorithm2: Testing Process**

**Input:** Testing data and the updated recognition system.

1) Feed the test data into the system, and get the fine-grained $M$-subcategory classification output $V_k, 1 \leq k \leq M$.

2) Sum over the output of step 1:

$$O_i = \sum_{k=1}^{M} W(i,k) * V_k, 1 \leq i \leq L, 1 \leq k \leq M \tag{7}$$

3) Select the maximum value in step 2 as the final $L$-category classification result, $R = \max(O_i), 1 \leq i \leq L$.

**Output:** The classification result.

# 3 The rule of determining cluster number $K_i$

The proper number $K_i$ of clusters for each category of our model SFM is determined by considering two aspects as follows:

**1) The overlapping feature distribution:**

The overlapping distribution of data from multiple categories in feature space is ubiquitous, especially when the features are not good enough to be well classified. The learning process with overlapping class distribution in such case will be difficult. Dividing the data in the feature space into multiple subcategories according to their distribution can alleviate the problem of overlapping feature distribution.

Thus, our first method to determine the proper number $K_i$ of subcategories is based on observing the t-Distributed Stochastic Neighbor Embedding (t-SNE) [13] 2-dimensional visualization of the high-dimensional feature space directly.

t-SNE has been proved to be an advanced technique for dimensionality reduction that is especially well suitable for the visualization of high-dimensional data by giving each data point a location in a two or three dimensional map.

Take the visualization result of feature distribution by t-SNE which is shown in Figure

1(a) as an example. Four classes are represented with four different symbols and colors. In this case, the subclasses of 2 in class 3 and 2 in class 4, i.e. $K_3 = 2$ and $K_4 = 2$ will be considered as the proper $K_i$ parameters in our model, and class 1 and class 2 will not be subdivided, i.e. $K_1 = 1$ and $K_2 = 1$.

**2) The class imbalance problem:**

The imbalanced class problem occurs when there are more samples in one class than other classes in a training dataset. Well balanced training samples are important [9].

Thus, our second method to determine the proper number $K_i$ of subcategories is based on the ratio of the number of each category samples to the number of the minority category samples. Subcategories of the nearly equal amount of samples are obtained after subdivision in each original category.

Suppose the training data $\{(x_1, y_1), \cdots (x_n, y_n)\}$ have, for example, 55 samples, which belong to three classes, labeled as 1, 2 and 3, i.e. $n = 55$, $y_i \in \mathbb{R}^3$, and there are 9, 29 and 17 samples in class 1, 2, and 3 respectively. In this case, the number of minority class samples is 9, the number of 29 and 17 are the triple and double of 9 approximately. Hence, the subclasses of 3 in class 2 and 2 in class 3, i.e. $K_2 = 3$ and $K_3 = 2$ will be considered as the proper $K_i$ parameters in our model, and class 1 will not be subdivided, i.e. $K_1 = 1$.

In this paper, we update the label for each sample of subclasses by the result of performing Sparse Subspace Clustering (SSC) [2] in each class.

# 4 Experiments

In this section, the experimental details and results based on our model SFM with Convolutional ISA Network [10], 3D CNN Network [6] and Action Bank features [19] are provided. Please note that, we are not aiming at proposing a new architecture to compete with others and achieve the state-of-the-art results. Our main contribution is that employing SFM, better performance than before can be achieved in the case of overlapping feature distribution, and the state-of-the-art results may be obtained.

## 4.1 Experiment on Hollywood2 dataset with Convolutional ISA

**Hollywood2 dataset.** The standard training set and testing set of Hollywood2 dataset [14] were used in our experiment.

**Details about experiment.** The resolution of all videos provided by Hollywood2 was set to 200×160, in consideration of the trade-off between training time and recognition performance. We implement SFM on the architecture of Convolutional ISA network [10], and the parameter settings were identical to [10]. 3000-dimensional feature vectors were obtained. For classification, softmax classifier was utilized in our experiment to conduct multi-classification, instead of using One-Against-All SVM with χ2 kernel in [10].

Utilizing SFM with our first method to determine the number of subcategories, by observing the t-SNE 2-dimensional visualization of the high-dimensional feature space of Hollywood2 (Figure 4(a)), we subdivided the 12 categories of actions into 25 subcategories approximately. Details about the subdivision are shown in Table 1.

Utilizing SFM with our second method to determine the number of subcategories, we subdivided the 12 categories of actions into 29 subcategories. After subdivision, the average number of the 29 subcategories samples was about 30, as shown in Table 1.

| Action classes Of Hollywood2 | Training samples | Subclasses (method 1) | Subclasses (method 2) | Action classes Of YouTube Action dataset | Subclasses (method 1) |
|---|---|---|---|---|---|
| Answering phone | 68 | 2 | 2 | Basketball shooting | 3 |
| Driving car | 85 | 4 | 3 | Biking | 3 |
| Eating | 40 | 2 | 1 | Diving | 2 |
| Fighting person | 54 | 2 | 2 | Golf swinging | 2 |
| Getting out of car | 51 | 2 | 2 | Horse riding | 3 |
| Hand shaking | 32 | 1 | 1 | Soccer juggling | 2 |
| Hugging person | 64 | 3 | 2 | Swinging | 4 |
| Kissing | 114 | 2 | 4 | Tennis swinging | 3 |
| Running | 135 | 1 | 4 | Trampoline jumping | 2 |
| Sitting down | 104 | 3 | 3 | Volleyball spiking | 2 |
| Sitting up | 24 | 1 | 1 | Walking with a dog | 3 |
| Standing up | 132 | 2 | 4 | | |

Table 1. Subdivision of Action Classes of Hollywood2

Table 3. Subdivision of Action Classes of YouTube Action dataset

| Method | SFM (method 1) | SFM (method 2) | Wang et al. [25] | Oneata et al. [16] | Jain et al. [5] | Mathe et al. [15] | Q. V. Le et al. [10] |
|---|---|---|---|---|---|---|---|
| Performance (mAP%) | **79.4%** | **74.0%** | 64.3% | 63.3% | 62.5% | 61.0% | **53.3%** |

Table 2. Comparison of our results on Hollywood2 to the results of other state-of-the-art approaches

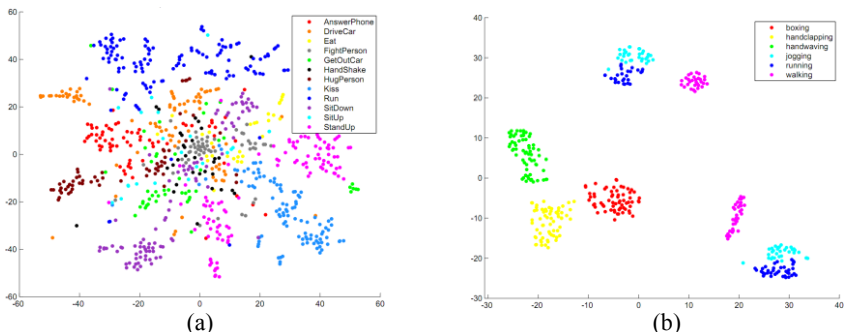

Figure 4: The visualization of ISA features by t-SNE [13]. (a) The 2-dimensional visualization of the high-dimensional feature space of Hollywood2. 12 action classes are represented with different colors. (b) The 2-dimensional visualization of the high-dimensional feature space of KTH. 6 action classes are represented with different colors. This figure is best viewed in color, zoomed in.

When performing the first method to set the number of subcategories, we applied SSC in each action category with the 2-dimensional feature vectors mapped from the original 3000-dimensional ones by t-SNE.

When performing the second method to determine the number of subcategories, SSC was employed in each action category with the original 3000-dimensional feature vectors.

**Experimental results.** The performance of our model on Hollywood2 dataset is reported in Table 2. The results of our work against the state-of-the-art reported results are shown. Accuracy (mAP) of 79.4% and 74.0% are obtained by our first and second method to determine the number of subcategories. Using SFM, the performance result on Hollywood2 has been remarkably improved and clearly outperforms the state-of-the-art result. Please note that, our model with ISA network improves the mAP from 53.3% to 79.4% and 74.0%, not from 64.3%. As comparisons, we equally group the samples of each action to $K_i$ subcategories, according to the same number $K_i$ of subcategories listed in Table 1, randomly grouping instead of using SSC, the performance is dropped to 48.5%. It suggests that the performance of SFM is partially benefit from a good clustering method.

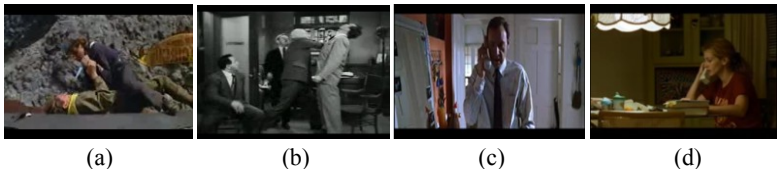

(a)         (b)         (c)         (d)

Figure 5: Sample frames of subcategories in Hollywood2. Using SFM, subcategories were divided meaningfully and presented obvious distinction. Sample frames of (a) *fighting person action, tumbling on the ground*, (b) *fighting person action, standing*, corresponding to the 2 subcategories divided from the original *fighting person* action category with method 1. Sample frames of (c) *answering phone action with front view*, (d) *answering phone action with side view*, corresponding to the 2 subcategories divided from the original *answering phone* action category with method 2.

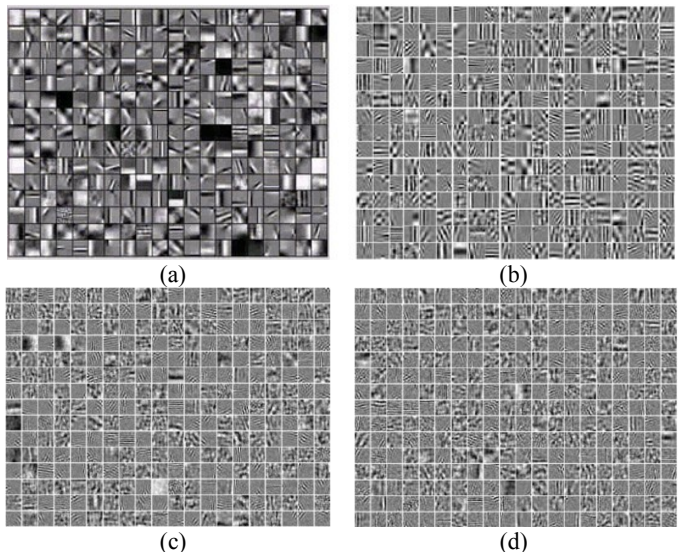

(a)         (b)

(c)         (d)

Figure 6: The first layer features visualization. The learned features are Gabor-like and Grating-like. (a) The first layer features from ISA [10]. (b) Our first layer features. Our model led to clearer and more distinctive features and better removal of the "dead" features. (c) The first layer features before subdivided (for example, all videos of *answering phone* action), with several "dead" features but clearer on the whole. (d) The first layer features of one subcategory (for example, the videos of *answering phone action with front view* which were subdivided from *answering phone* action), "dead" features were almost removed but less clear, due to the less training video samples of the subcategory. Best zoomed in.

With our two methods to determine the number of subcategories, sample frames of subcategories are shown in Figure 5. As it can be observed, with our model, those subcategories are divided meaningfully and present obvious distinction.

**Feature visualization.** Features learned for the first layer can be inspected in Figure 6. The learned features are Gabor-like and Grating-like. Compared with the first layer features from ISA [10] (Figure 6(a)), our model resulted in cleaner and more distinctive features with significant removal of the "dead" features (Figure 6(b)). Comparison of the first layer features between original category and subcategory is presented in Figure 6(c) and Figure 6(d). The original category had several "dead" features but was clearer on the whole. On the contrary, the subcategory had less "dead" features but was less clear. This was due to the fact that the number of training videos for each subcategory was decreased.

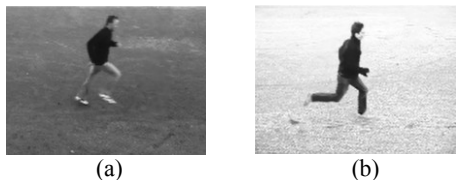

(a) (b)

Figure 7: Sample frames of subcategories of *running* action with method 1 from KTH, the meaningful difference of scenes can be inspected

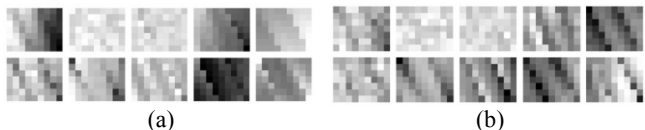

(a) (b)

Figure 8: (a) The first layer features before subdivision. (b) The first layer features after subdivision. The first column is the gray channel information. The second and the third column are the gradient value in horizontal and vertical direction. The last two columns are the optical flow in horizontal and vertical direction. Best zoomed in.

| Method | SFM (method 1) | Wang et al. [26] | Q. V. Le et al. [10] | Sun et al. [22] | Ji et al. [6] |
|---|---|---|---|---|---|
| Accuracy | **94.0%** | 94.2% | 93.9% | 93.1% | **90.2%** |

Table 4. Comparison of our results on KTH to the results of other state-of-the-art methods

## 4.2 Experiment on YouTube Action dataset with Convolutional ISA

**YouTube Action dataset.** As identical to the paper of ISA [10], Leave-One-Out Cross-Validation was used in our experiment on the YouTube Action dataset [12].

**Details about experiment.** The parameter settings in our experiments were identical to the description in [10]. 3000-dimensional feature vectors were obtained.

The class imbalance problem of YouTube Action dataset was not obvious. The maximum number of class samples was 199 and the minimum one was 116, other numbers had a little differences. Thus, we considered the first method to determine the number of subcategories, 11 categories were subdivided into 29 subcategories approximately(Table 3).

**Experimental result.** Employing SFM with method 1, the performance result on YouTube Action dataset has been considerably improved from 75.8% [10] to 82.5%.

## 4.3 Experiment on KTH dataset with 3D CNN

**KTH dataset.** As identical to [6], the same training set and testing set of the KTH dataset [20] were used in our experiment.

**Details about experiment.** The class imbalance problem of KTH dataset didn't exist (number of each training class samples was almost the same), our first method to determine the number of subcategories was used. By observing the t-SNE 2-dimensional visualization of the high-dimensional feature space of KTH (Figure 4(b)), 6 action classes were subdivided into 9 subcategories. We obtained 2 subcategories of *jogging*, 2 subcategories of *running* and 2 subcategories of *walking*. The others were not subdivided.

We performed SSC in each class with the 2-dimensional feature vectors mapped from the original 128-dimensional ones by t-SNE, and then updated the label of each sample. The network was trained again with the updated labels.

**Experimental result.** Table 4 shows the comparison of our result with other recently proposed methods which are reported as the state-of-the-art in the last few years. Our model improved the accuracy with 3D CNN from 90.2% [6] to 94.0%.

Figure 7 shows the result after subdivision for the action of *running*. Utilizing SFM, the meaningful difference of scenes can be inspected in Figure 7.

**Feature visualization.** Figure 8 shows the feature visualization result of the first layer. The gray channel feature from the original category is better than after subdivision. The features in Figure 8(b) are vaguer. The reason is that after subdivision, the samples in each category are halved. For the gradient and optical flow features, Figure 8(b) is clearer than Figure 8(a) which indicates that after subdivision, better features are learned.

## 4.4 Experiment on UCF50 dataset with Action bank features

**UCF50 Dataset.** Specific details about the large-scale UCF50 dataset can be found in [18].

**Details about experiment.** Available precomputed UCF50 [18] Action Bank [19] 14965-dimensional feature vectors could be found from [29]. We conducted UCF50 through 10-fold video-wise cross-validation, which is identical to the paper of Action Bank [19].

The t-SNE 2-dimensional visualization of the high-dimensional feature vectors of UCF50 was not distinct to be clearly observed due to both the relatively large number of action classes (50 classes) and the relatively low accuracy (76.4% reported in [19]), our second method to determine the number of subcategories was used. With precomputed UCF50 Action Bank feature vectors, 50 categories were subdivided into 124 subcategories.

**Experimental result.** Employing SFM with method 2 based on precomputed UCF50 Action Bank feature vectors, the performance result on UCF50 dataset has been slightly improved from 76.4% [19] to 76.9%.

## 5 Conclusion

In most recognition tasks, it is a common phenomenon that features of multiple categories are overlapping distributed, which leads to overfitting. We proposed a novel Subdivision-Fusion Model to improve the human action recognition accuracy. First, we converted the samples to the feature vectors by a feature extractor, for example a deep network. Then we clustered each category into multiple subcategories. Two methods to determine the proper number of clusters were provided. The first method was directly observing the feature distribution of the high-dimensional feature space by t-SNE 2-dimensional visualization. The second method was based on the ratios of each category to the minimum number of samples. The labels were updated according to the clustering results of SSC. At the fusion stage, we combined one more layer to the classifier to convert the multi-subcategories classification back to the original classification problem. We evaluated our model SFM on four popular datasets to demonstrate the enhancement in performance.